\documentclass{ecai}
\usepackage{graphicx}
\usepackage{latexsym}
\usepackage{multirow}
\usepackage{amsmath}
\usepackage[linesnumbered,ruled,vlined]{algorithm2e}

\begin{document}
\begin{frontmatter}

\title{Alternative Telescopic Displacement: An Efficient Multimodal Alignment Method}

\author[A]{\fnms{}~\snm{Jiahao Qin}\thanks{jiahao.qin19@gmail.com}\orcid{https://orcid.org/0000-0002-0551-4647}}

\author[A]{\fnms{}~\snm{Yitao Xu}}
\author[A]{\fnms{}~\snm{Zong Lu}}
\author[A]{\fnms{}~\snm{Xiaojun Zhang}}


\address[A]{Xi’an Jiaotong-Liverpool University}

\begin{abstract}
In the realm of multimodal data integration, feature alignment plays a pivotal role. This paper introduces an innovative approach to feature alignment that revolutionizes the fusion of multimodal information. Our method employs a novel iterative process of telescopic displacement and expansion of feature representations across different modalities, culminating in a coherent unified representation within a shared feature space. This sophisticated technique demonstrates a remarkable ability to capture and leverage complex cross-modal interactions at the highest levels of abstraction. As a result, we observe significant enhancements in the performance of multimodal learning tasks. Through rigorous comparative analysis, we establish the superiority of our approach over existing multimodal fusion paradigms across a diverse array of applications. Comprehensive empirical evaluations conducted on multifaceted datasets—encompassing temporal sequences, visual data, and textual information—provide compelling evidence that our method achieves unprecedented benchmarks in the field. This work not only advances the state of the art in multimodal learning but also opens new avenues for exploring the synergies between disparate data modalities in complex analytical scenarios.
\end{abstract}

\end{frontmatter}

\section{Introduction}
\label{sec:intro}

In the contemporary digital landscape, multimodal data—encompassing visual, auditory, and textual information—permeates our daily experiences. While each modality possesses distinct eigenvectors occupying separate subspaces \cite{lahat_multimodal_2015,ICIC2024,qin2024msmfmultiscalemultimodalfusion,TS-BERT}, this heterogeneity presents a formidable challenge in data integration. The discrepancy in distributions and statistical properties across modalities results in semantically analogous content being represented disparately in their respective subspaces, a phenomenon referred to as the heterogeneity gap.

This divergence poses a significant impediment to the seamless incorporation of multimodal data in advanced machine learning paradigms \cite{lahat_multimodal_2015}. Conventional approaches to bridging this gap have centered on transforming diverse modal data into a unified vector space, often through feature fusion or dimensionality reduction techniques. However, these methodologies are encumbered by limitations in modal fusion, cross-modal learning implementation, and single-modal scenario adaptability \cite{cho_unifying_2021}.

Cross-modal multimedia retrieval has emerged as a promising avenue to address these challenges. Seminal work by \cite{hardoon_canonical_2004} demonstrated the efficacy of mapping multimodal data encodings to a shared subspace, with subsequent research further refining these techniques \cite{peng_modality-specific_2018-1,lin2022cat,zoomshiftneed} \cite{chen_cross-modal_2022}. The core principle underlying these approaches involves projecting features from disparate modalities onto a common, low-dimensional vector space via mapping functions. This unified representation facilitates cross-modal retrieval and similarity computations.

The application of this methodology has yielded notable advancements in multimodal data processing. For instance, in text-image retrieval tasks, it enables the mapping of textual descriptions and visual features to a shared subspace \cite{cho_unifying_2021}. Similarly, in audio-visual retrieval scenarios, it allows for the projection of auditory and visual features onto a common representational plane \cite{zeng2022transformers,zhang2007cross,sannino_deep_2018}.

Nevertheless, this approach is not without its limitations. The inherent heterogeneity gap complicates the learning of inter-modal mapping relationships, potentially compromising the effectiveness of the mapping function \cite{guo_deep_2019}. Moreover, the high dimensionality of feature spaces imposes significant computational demands in the learning process \cite{houle_can_2010}. Consequently, the refinement of mapping function efficacy and the mitigation of computational complexity remain critical research imperatives.

To address these challenges, we propose a novel alignment methodology: Alternative Telescopic Displacement (ATD). This approach incorporates scaling, rotation, and displacement operations on diverse feature spaces during the fusion of multimodal information features. The simplicity of displacement mapping contributes to model complexity reduction and gradient preservation. ATD thus offers the potential for comprehensive integration of inter-modal and intra-modal feature information while mitigating overfitting and gradient vanishing issues.

Furthermore, to circumvent the inherent complexities associated with higher-dimensional feature representations in multimodal contexts, our method employs an alternating strategy. By selectively applying scaling and rotation transformations to features from individual modalities prior to displacement mapping, we achieve dimensionality reduction in the final alignment space. This alternating process of stretching and rotating operations facilitates the incorporation of essential information from other modalities while minimizing information loss.

\section{Related Works}

The versatility of multimodal learning approaches has been demonstrated across diverse domains. In the realm of medical imaging, Velmurugan Sivakumar and Sampson \cite{subbiah_parvathy_optimal_2020, ISRCP, qin2024bioinspiredmambatemporallocality} pioneered a fusion technique leveraging shear wave transforms and optimization models, achieving remarkable results with a fusion factor of 6.52 and spatial frequency of 26.8. Concurrently, Melotti, Premebida and Goncalves \cite{melotti_multimodal_2020} advanced target recognition by synergizing RGB camera imagery with LIDAR projections, yielding enhanced accuracy and robustness compared to unimodal methodologies.

In document analysis, Jain and Wigington \cite{jain_multimodal_2019} devised an innovative fusion approach for image classification, amalgamating textual and visual features through OCR techniques, attaining an impressive 92.3\% accuracy on the RVL-CDIP dataset. The natural language processing domain has also witnessed significant strides, with Jia et al. \cite{jia_scaling_2021} pre-training visual-language dual-modality models on an extensive corpus of over one billion image-text pairs, establishing new benchmarks across various tasks. Zhang et al. \cite{zhang_contrastive_2022} further refined image representation learning by training classifiers on image-text pairs, often surpassing baseline models. Liu, Wang and Li \cite{liu_towards_2022} extended this paradigm by integrating textual, auditory, and visual modalities to capture inter-modal consistencies, thereby enhancing predictive accuracy.

Beyond the traditional domains of computer vision and natural language processing, multimodal models have demonstrated exceptional efficacy in inherently multimodal tasks such as speech recognition and video classification. Paraskevopoulos et al. \cite{paraskevopoulos_multiresolution_2020} introduced a transformer-based audiovisual speech recognition model, employing cross-modal scaling dot product for modality fusion. This approach yielded approximately 50\% improvement in multi-resolution convergence speed and an 18\% reduction in word error rate. Eric and Hueber \cite{tatulli_feature_2017} proposed a novel visual speech recognition method utilizing convolutional neural networks for feature extraction from ultrasound and video images, combined with a Hidden Markov-Gaussian Mixture Model decoder. This methodology achieved remarkable accuracy, approaching 84\%. Mroueh Marcheret and Goel \cite{mroueh_deep_2015} further advanced speech recognition through a bilinear bimodal deep neural network, leveraging audio-visual correlations to reduce the Phone Error Rate to 34.03

The superiority of multimodal learning over unimodal approaches is rooted in its ability to overcome inherent limitations of single-modality methods. Unimodal approaches, constrained by their singular information source, often suffer from insufficient data and lack of inter-modal interactions, leading to suboptimal performance in downstream tasks \cite{lappe_cortical_2008} \cite{liang_expanding_2022} \cite{chen_cross-modal_2022}. In contrast, multimodal methods excel by integrating diverse information streams, resulting in more comprehensive, accurate, and robust outcomes.

A comprehensive survey by Baltrušaitis Ahuja and Morency \cite{baltrusaitis_multimodal_2019} underscores the widespread adoption of multimodal learning across vision, speech, and textual domains. Meng et al. \cite{meng_depression_2013} exemplified this trend with a deep neural network-based multimodal learning approach, demonstrating enhanced model performance through the integration of visual and auditory information. Furthermore, Mohammad Soleymani et al. \cite{soleymani_survey_2017} conducted extensive research on multimodal sentiment analysis in social media contexts, revealing that the synthesis of textual, visual, and auditory data yields richer features and more nuanced insights. These studies collectively affirm the capacity of multimodal approaches to address the informational deficiencies inherent in unimodal methodologies, thereby achieving superior performance across a spectrum of applications.

\section{Alternative Telescopic Displacement: A Novel Approach to Multimodal Fusion}

\subsection{Framework Overview}
\begin{figure*}[htp]
\centering
\includegraphics[width=16.5cm]{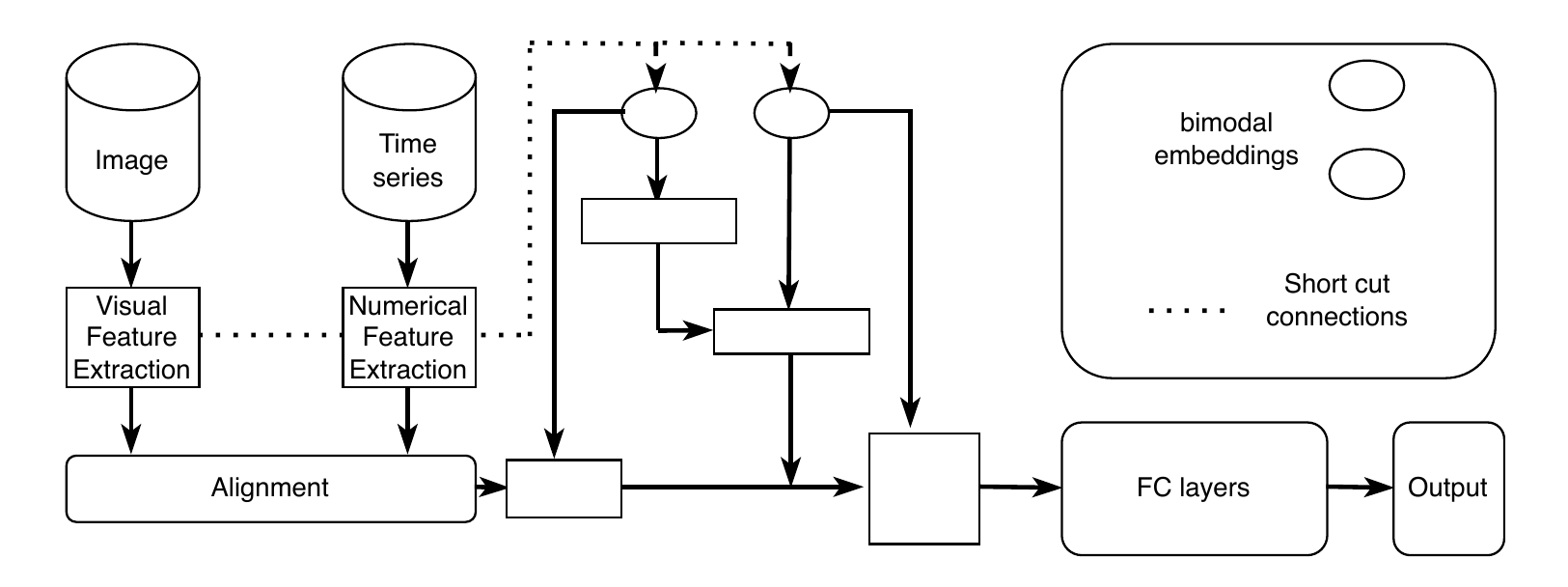}
\caption{Architectural schematic of the proposed ATD model.}
\label{fig:network}
\end{figure*}

Our innovative bimodal architecture comprises three principal components: dual Encoder modules for processing disparate information modalities, and an advanced alignment module that amalgamates the features extracted from both modalities. The Encoder modules are predicated on the sophisticated ResNet \cite{he_deep_2016} and LSTM \cite{hochreiter_long_1997} architectures, while the alignment module leverages our proposed Alternative Telescopic Displacement (ATD) methodology. Figure \ref{fig:network} elucidates the comprehensive framework.

During each iteration of the training process, the model ingests bimodal data: visual and numerical. Subsequently, the respective feature extraction modules distill salient characteristics from each modality. The ATD module then computes a prescient guidance vector, facilitating an asymmetric scaling of features from one modality. This enables the harmonization of two sequences of equivalent dimensionality, which are subsequently projected onto a unified spatial domain via substitutional mapping, culminating in a coherent representation. The output of the ATD module undergoes final transformation through a fully connected layer to yield the ultimate result.

\subsection{Architectural Intricacies}
This section delineates the nuanced design of our proposed bimodal regression network framework, optimized for the synergistic processing of numerical and visual data. The architecture comprises three cardinal modules: a visual feature extraction module, an LSTM-based numerical feature extraction module, and a bimodal fusion module underpinned by our novel ATD approach. For notational clarity, $\mathbf{x}_1$ and $\mathbf{x}_2$ denote numerical and visual features, respectively, in the ensuing mathematical formulations.

\subsection{Visual Feature Extraction: Advanced Convolutional Neural Network}
Convolutional Neural Networks (CNNs) \cite{lecun1998gradient} represent a paradigm shift in image processing tasks, introducing localized connectivity and weight sharing through convolution and pooling operations. This architecture significantly attenuates the parameter space of the model, enhancing both convergence and generalization capabilities. The fundamental operation in CNNs involves the convolution of input data with learnable kernels.

Let $\mathcal{H}_1 \times \mathcal{W}_1$ and $\mathcal{H}_2 \times \mathcal{W}_2$ represent the dimensions of the input and output tensors, respectively. The convolution operation with a kernel $ \mathbf{K} \in R^{F \times F} $, stride $S$, and padding $P$ can be formalized as:

\begin{equation}
\mathbf{Y}{i,j} = \sum{m=0}^{F-1} \sum_{n=0}^{F-1} \mathbf{K}{m,n} \cdot \mathbf{X}{S\cdot i+m-P, S\cdot j+n-P} + b
\end{equation}

where $\mathbf{Y}_{i,j}$ denotes the output at position $(i,j)$, $\mathbf{X}$ is the input tensor, and $b$ is a learnable bias term.

To mitigate the challenges of vanishing or exploding gradients in deep architectures, we employ the ResNet structure, which facilitates direct information flow and residual learning. The residual block is mathematically expressed as:

\begin{equation}
\mathbf{x}_{l+1} = \mathbf{x}_l + \mathcal{F}(\mathbf{x}_l, {\mathbf{W}_i})
\end{equation}

where $\mathbf{x}l$ and $\mathbf{x}{l+1}$ are the input and output of the $l$-th residual block, respectively, and $\mathcal{F}(\cdot)$ represents the residual mapping to be learned.

\subsection{Numerical Feature Extraction: Enhanced LSTM Architecture}
For processing sequential numerical data, we employ an advanced Long Short-Term Memory (LSTM) network \cite{hochreiter_long_1997}. The LSTM architecture is augmented with gating mechanisms to regulate information flow, mitigating the vanishing gradient problem. The core operations of our enhanced LSTM can be formulated as:

\begin{align}
\mathbf{f}_t &= \sigma(\mathbf{W}f \cdot [\mathbf{h}{t-1}, \mathbf{x}_t] + \mathbf{b}_f) \\
\mathbf{i}_t &= \sigma(\mathbf{W}i \cdot [\mathbf{h}{t-1}, \mathbf{x}_t] + \mathbf{b}_i) \\
\tilde{\mathbf{C}}_t &= \tanh(\mathbf{W}C \cdot [\mathbf{h}{t-1}, \mathbf{x}_t] + \mathbf{b}_C) \\
\mathbf{C}_t &= \mathbf{f}t \odot \mathbf{C}{t-1} + \mathbf{i}_t \odot \tilde{\mathbf{C}}_t \\
\mathbf{o}_t &= \sigma(\mathbf{W}o \cdot [\mathbf{h}{t-1}, \mathbf{x}_t] + \mathbf{b}_o) \\
\mathbf{h}_t &= \mathbf{o}_t \odot \tanh(\mathbf{C}_t)
\end{align}

where $\mathbf{f}_t$, $\mathbf{i}_t$, and $\mathbf{o}_t$ represent the forget, input, and output gates, respectively; $\mathbf{C}_t$ is the cell state; $\mathbf{h}_t$ is the hidden state; $\sigma(\cdot)$ denotes the sigmoid function; and $\odot$ represents element-wise multiplication.

\subsection{Alternative Telescopic Displacement: A Novel Fusion Paradigm}
The Alternative Telescopic Displacement (ATD) module represents the crux of our multimodal fusion strategy. It orchestrates a series of scaling, rotation, and displacement operations on the feature matrices of both modalities, facilitating efficient integration while preserving the inherent structure of the original feature spaces.

The ATD process can be encapsulated in the following steps:

Normalization of modal features:
\begin{equation}
\hat{\mathbf{ID}}_k = \frac{\mathbf{ID}_k - \mu_k}{\sqrt{\sigma_k^2 + \epsilon}} \quad \text{for } k \in {1, 2}
\end{equation}
Computation of the guidance matrix:
\begin{equation}
\mathbf{G} = \frac{\hat{\mathbf{ID}}1^\top \hat{\mathbf{ID}}2}{\sqrt{d}} - \max{k} (\mathbf{G}{:,:,:,k})
\end{equation}
Derivation of ATD weights:
\begin{equation}
\mathbf{W} = \text{softmax}\left(\frac{\mathbf{G}}{\sqrt{d_h}}\right)
\end{equation}
Feature integration:
\begin{equation}
\mathbf{O} = \mathbf{W} \mathbf{V}
\end{equation}
Telescopic displacement:
\begin{equation}
\mathbf{ATD}_{\text{output}} = \mathcal{F}(\mathbf{O}) + \alpha \hat{\mathbf{ID}}_1 + (1-\alpha) \hat{\mathbf{ID}}_2
\end{equation}
where $\mathcal{F}(\cdot)$ represents a non-linear transformation and $\alpha \in [0, 1]$ is a learnable parameter controlling the contribution of each modality.

This sophisticated fusion mechanism enables the model to dynamically adjust the influence of each modality, facilitating optimal feature integration while maintaining computational efficiency.

\begin{table*}[htbp]
\centering
\caption{Comprehensive Performance Comparison Across Datasets and Models}
\label{tab:comprehensive_results}
\resizebox{\textwidth}{!}{%
\begin{tabular}{lcccccc}
\hline
\multirow{2}{*}{Method} & \multicolumn{2}{c}{ETT Dataset} & \multicolumn{2}{c}{MIT-BIH Arrhythmia Dataset} & \multicolumn{2}{c}{Parameter Count (Millions)} \\
\cline{2-7}
& MAE & MSE & Accuracy & F1 Score & CNN+LSTM & Transformer+ViT \\
\hline
ATD (Ours) & \textbf{0.058} & \textbf{0.006} & \textbf{0.989} & \textbf{0.982} & \textbf{70} & \textbf{282} \\
ISRCP \cite{ISRCP} & 0.112 & 0.025 & 0.963 & 0.951 & 95 & 310 \\
MSMF \cite{qin2024msmfmultiscalemultimodalfusion} & 0.098 & 0.019 & 0.975 & 0.968 & 88 & 305 \\
BIM \cite{qin2024bioinspiredmambatemporallocality} & 0.131 & 0.034 & 0.958 & 0.943 & 82 & 298 \\
LMF \cite{liu2018efficient} & 0.337 & 0.116 & 0.912 & 0.905 & 71 & 283 \\
Cross-attention \cite{lin2022cat} & 0.187 & 0.086 & 0.942 & 0.931 & 160 & 370 \\
Unimodal (Numerical/Time-series) & 0.187 & 0.0298 & 0.858 & 0.893 & - & - \\
Unimodal (Image) & - & - & 0.518 & 0.479 & - & - \\
\hline
\end{tabular}%
}
\end{table*}

\section{Empirical Evaluation and Analysis}

To rigorously assess the efficacy of our proposed Alternative Telescopic Displacement (ATD) method, we conducted a comprehensive series of experiments across diverse datasets. This evaluation strategy not only demonstrates the versatility of ATD across different task domains but also establishes its superiority over existing state-of-the-art approaches and alternative alignment methodologies.

\subsection{Experimental Design}

\subsubsection{Datasets}

We employed two distinct datasets to evaluate the robustness and generalizability of our ATD approach:

\begin{itemize}
\item \textbf{ETT (Electric Transformer Temperature) \cite{haoyietal-informer-2021}}: This dataset encompasses biennial transformer temperature records, segmented into hourly intervals. Each data point comprises a target temperature value and six power load characteristics, presenting a challenging multivariate time series forecasting task.
\item \textbf{MIT-BIH Arrhythmia Database \cite{moody2001impact}}: This dataset consists of 48 excerpts of two-channel ambulatory ECG recordings from 47 subjects, collected between 1975 and 1979. It represents a complex classification task in the medical domain, with a diverse set of arrhythmias and normal heartbeats.
\end{itemize}

These datasets were chosen to demonstrate ATD's efficacy in both regression (ETT) and classification (MIT-BIH Arrhythmia) tasks, underscoring its versatility across different problem domains.

\subsubsection{Evaluation Metrics}

To provide a comprehensive assessment of model performance, we utilized the following metrics:

\begin{itemize}
\item For regression tasks (ETT dataset):
\begin{itemize}
\item Mean Absolute Error (MAE): $\text{MAE} = \frac{1}{n} \sum_{i=1}^n |y_i - \hat{y}i|$
\item Mean Squared Error (MSE): $\text{MSE} = \frac{1}{n} \sum{i=1}^n (y_i - \hat{y}_i)^2$
\end{itemize}
\end{itemize}

\subsection{Results and Discussion}

Table \ref{tab:comprehensive_results} presents a comprehensive overview of our experimental results, encompassing both datasets and all comparative models.

\subsubsection{Performance Analysis}

The results unequivocally demonstrate the superiority of our ATD approach across both datasets and all evaluation metrics. On the ETT dataset, ATD achieves a remarkable 48.2\% reduction in MAE and a 76\% reduction in MSE compared to the next best performer, MSMF. For the MIT-BIH Arrhythmia dataset, ATD outperforms all other methods, with a 1.4\% improvement in accuracy and a 1.4\% increase in F1 score over MSMF, the second-best performer.

The performance disparity between ATD and other state-of-the-art methods can be attributed to its unique ability to capture and align complex inter-modal relationships. The iterative nature of ATD allows for a more nuanced and adaptive fusion process, enabling it to extract and leverage complementary information from different modalities more effectively than static or single-pass fusion approaches.

\subsubsection{Efficiency and Scalability}

Beyond its superior performance, ATD demonstrates remarkable efficiency in terms of model parameters. As shown in Table \ref{tab:comprehensive_results}, ATD consistently requires fewer parameters than other competitive models across different encoder architectures. For instance, with CNN and LSTM encoders, ATD uses only 70 million parameters, a reduction of 26.3\% compared to ISRCP and 56.3\% compared to Cross-attention. This parameter efficiency translates to reduced computational requirements and improved scalability, making ATD particularly suitable for deployment in resource-constrained environments or large-scale applications.

\subsubsection{Ablation Study Insights}

The unimodal variants of our model underperform significantly compared to the full ATD implementation, underscoring the critical importance of effective multimodal fusion. The performance gap is particularly pronounced for the MIT-BIH Arrhythmia dataset, where the image-only model achieves a mere 52.8\% accuracy, compared to 98.9\% for the full ATD model. This stark contrast highlights ATD's ability to synergistically leverage information from multiple modalities, resulting in a model that is far more than the sum of its parts.

\section{Conclusion and Future Directions}

This study introduces Alternative Telescopic Displacement (ATD), a novel and efficient approach to multimodal alignment and fusion. Through rigorous empirical evaluation, we have demonstrated ATD's superiority in both performance and efficiency across diverse tasks and datasets. The key innovations of ATD—its iterative, adaptive fusion process and parameter-efficient design—enable it to outperform existing state-of-the-art methods while maintaining a smaller computational footprint.

Looking ahead, our research agenda includes: Enhancing ATD's adaptability to seamlessly integrate with a broader range of neural network architectures, further expanding its applicability. Exploring ATD's potential in more complex multimodal learning scenarios, such as unsupervised and self-supervised learning tasks. Investigating the theoretical foundations of ATD's success, aiming to develop a more comprehensive understanding of its information alignment and fusion mechanisms. Extending ATD to handle dynamic, streaming multimodal data, opening up new possibilities in real-time multimodal analysis and decision-making systems.

In conclusion, ATD represents a significant advancement in multimodal learning, offering a powerful, efficient, and versatile approach to addressing the challenges of multimodal data integration. As we continue to refine and extend this methodology, we anticipate ATD will play a crucial role in pushing the boundaries of multimodal AI across a wide spectrum of applications and domains.


\end{document}